\documentclass[twocolumn]{NobArticle}
\usepackage{algorithm}
\usepackage{algorithmic}
\usepackage{tabularx}
\usepackage[table]{xcolor}
\usepackage{tabu}
\runninghead{Shortened Running Article Title}
\footertext{\textit{Journal X} (2023) 12:684}

\title{EfficientMonoHair: Fast Strand-Level Reconstruction from Monocular Video via Multi-View Direction Fusion}

\author{
Da Li \quad Dominik Engel \quad Deng Luo \quad Ivan Viola\\
King Abdullah University of Science and Technology\\
Saudi Arabia\\
{\tt\small \{da.li, dominik.engel, deng.luo, ivan.viola\}@kaust.edu.sa}
}

\date{
}

\renewcommand{\maketitlehookd}{%
\begin{abstract}
Strand-level hair geometry reconstruction is a fundamental problem in virtual human modeling and the digitization of hairstyles. However, existing methods still suffer from a significant trade-off between accuracy and efficiency. Implicit neural representations can capture the global hair shape but often fail to preserve fine-grained strand details, while explicit optimization-based approaches achieve high-fidelity reconstructions at the cost of heavy computation and poor scalability.
To address this issue, we propose EfficientMonoHair, a fast and accurate framework that combines the implicit neural network with multi-view geometric fusion for strand-level reconstruction from monocular video. Our method introduces a fusion-patch-based multi-view optimization that reduces the number of optimization iterations for point cloud direction, as well as a novel parallel hair-growing strategy that relaxes voxel occupancy constraints, allowing large-scale strand tracing to remain stable and robust even under inaccurate or noisy orientation fields.
Extensive experiments on representative real-world hairstyles demonstrate that our method can robustly reconstruct high-fidelity strand geometries with accuracy. On synthetic benchmarks, our method achieves reconstruction quality comparable to state-of-the-art methods, while improving runtime efficiency by nearly an order of magnitude.
    \medskip

    \small{\textbf{Index Terms:} Strand-Level Reconstruction, Monocular video, Multi-View Direction Fusion.}
\end{abstract}
}

\begin{document}

\small
\maketitle

\vspace{-6mm}
\section{Introduction}

\begin{figure}
    \centering
    \includegraphics[width=\linewidth]{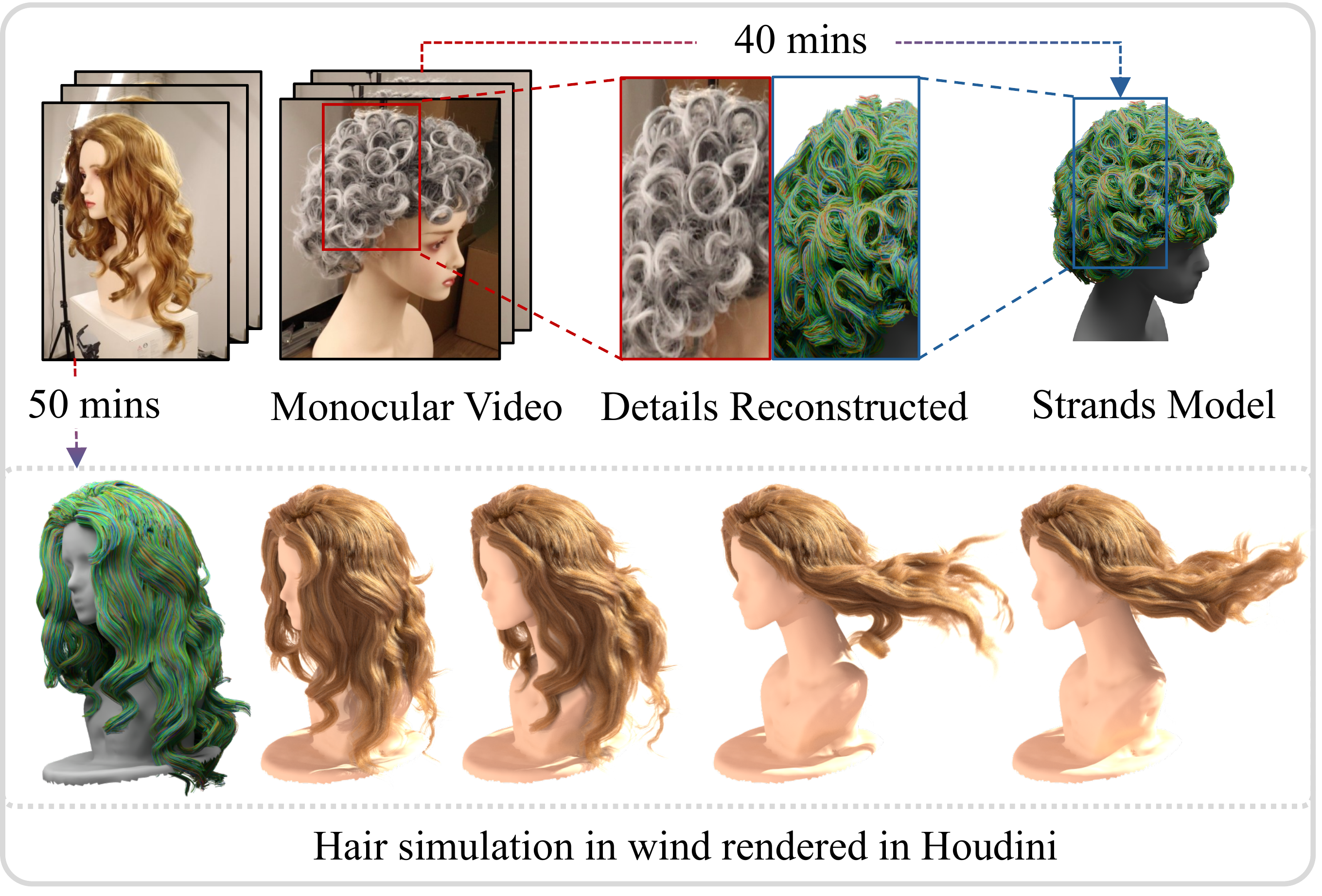}
    \caption{Our method efficiently reconstructs strand-level hair geometry from monocular video input. The top part demonstrates that EfficientMonoHair accurately captures diverse hairstyles, including curly hair, faithfully preserving both the global silhouette and fine local curls. The bottom part illustrates the scalp-attached strand reconstruction, which can be seamlessly imported into existing graphics systems for photorealistic rendering and physically based hair simulation. The colorful strands are rendered in Blender~\cite{blender2025} and the hair simulation is conducted in Houdini~\cite{houdini2024}.}
    \label{fig:teaser}
    \vspace{-4mm}
\end{figure}

Accurate modeling of individual hair strands is a crucial component in achieving high-fidelity digital humans and photorealistic virtual avatars. Unlike facial~\cite{FLAME:SiggraphAsia2017} or body geometry~\cite{loper17SMPLSkinnedMultiperson2015,fengCollaborativeRegressionExpressive2021}, which often follows standardized and relatively rigid representations, hair exhibits highly non-rigid behavior and complex topological structures. These characteristics make hair geometry reconstruction one of the long-standing challenges in the digitization of virtual humans.
Recent digital human systems, leveraging neural radiance fields (NeRF)~\cite{mildenhall1NeRFRepresentingScenes2022} or 3D Gaussian splatting (3DGS)~\cite{kerbl3DGaussianSplatting2023, wangMeGAHybridMeshGaussian2024}, have demonstrated impressive realism in rendering overall hair appearance. However, such approaches primarily recover the volumetric shape of hair~\cite{tangGAFGaussianAvatar2025,ramonH3DNetFewShotHighFidelity2021,zhouHeadStudioTextAnimatable2024,liaoHHAvatarGaussianHead2024,zhengAvatarImplicitMorphable2022,fengLearningDisentangledAvatars2023,wangMeGAHybridMeshGaussian2024,giebenhainMonoNPHMDynamicHead2024,grassalNeuralHeadAvatars2022,zhengPointAvatarDeformablePointbased2023}, failing to capture strand-level spatial details that are crucial for physically plausible hair motion, detailed appearance modeling, and realistic animation, like the simulation shown in Figure~\ref{fig:teaser}.

Early hair reconstruction approaches were predominantly based on explicit geometric optimization. These methods~\cite{paris2004capture,luoStructureawareHairCapture2013,luoMultiviewHairCapture2012,luoWideBaselineHairCapture2013} relied on multi-view imagery and handcrafted geometric constraints to reconstruct strand structures; however, they suffer from extremely low computational efficiency and calibration overhead.
With the advent of deep learning, implicit neural representation–based methods learn implicit shape and orientation distributions of hair~\cite{zhouHairNetSingleViewHair2018,saito3DHairSynthesis2018,wuNeuralHDHairAutomaticHighfidelity2022,rosuNeuralStrandsLearning2022}. Although they have achieved remarkable improvements in automation and reconstruction speed, they often struggle to recover fine-grained directional details in complex hairstyles—particularly in cases involving curly hair or intersecting strand structures.
Recent research trends have shifted toward hybrid explicit–implicit frameworks, aiming to combine the interpretability of explicit geometry with the generalization of implicit neural representations.
These methods leverage Gaussian fields~\cite{zakharovHumanHairReconstruction2024,luoGaussianHairHairModeling2024} and diffusion models~\cite{sklyarovaNeuralHaircutPriorGuided2023} to enhance the overall fidelity of hair geometry and appearance.
However, they still face challenges in time-consuming directional field optimization and inconsistent strand-level details across views. In particular, hair reconstruction methods based on CT scanning are highly limited in practical use~\cite{shen2023ct2hair}.

Achieving a balance between reconstruction accuracy, geometric detail, and computational efficiency remains a key open problem in strand-level hair modeling. Although MonoHair~\cite{wuMonoHairHighFidelityHair2024} has demonstrated substantial improvements over previous state-of-the-art approaches, such as NeuralHaircut~\cite{sklyarovaNeuralHaircutPriorGuided2023} in both runtime and hairstyle diversity, it still requires 4–9 hours to achieve high-quality strand reconstruction for a single hairstyle, which is significantly slower than typical multi-view avatar reconstruction pipelines~\cite{tangGAFGaussianAvatar2025,ramonH3DNetFewShotHighFidelity2021,zhouHeadStudioTextAnimatable2024,liaoHHAvatarGaussianHead2024,zhengAvatarImplicitMorphable2022,fengLearningDisentangledAvatars2023,wangMeGAHybridMeshGaussian2024,giebenhainMonoNPHMDynamicHead2024,grassalNeuralHeadAvatars2022,zhengPointAvatarDeformablePointbased2023}, which usually complete within 30–60 minutes. Therefore, further improving the reconstruction efficiency without compromising strand-level fidelity remains an essential yet unresolved challenge.

To address the aforementioned challenges, we present EfficientMonoHair, an efficient framework for strand-level hair reconstruction from monocular video. Our approach introduces a Fusion-Patch-based Multi-View Optimization (\textbf{FPMVO}) mechanism that enables fast directional consistency optimization for the external strand orientation field, while leveraging the generalization ability of implicit neural inference to estimate the internal orientation field. Specifically, FPMVO establishes patch-level consistency constraints across multiple views, allowing robust aggregation and refinement of directional information. This significantly reduces the number of optimization iterations while maintaining global stability of the external orientation field. In addition, we design a Parallel Hair Growing (\textbf{PHG}) strategy that simultaneously generates a large number of hair strands within a unified occupancy–orientation volume. This GPU-parallelized growing process is very robust to inconsistencies in the incoming orientation volumes and ensures both geometric fidelity and curvature continuity by relaxing voxel overlapping occupancy constraints. As a result, it enables substantial improvements in reconstruction quality, as well as efficiency. Extensive experiments on both real and synthetic hairstyles demonstrate that EfficientMonoHair robustly reconstructs high-fidelity strand geometries and achieves significant runtime advantages.

Compared with state-of-the-art (SOTA) methods, our approach offers the following main contributions:

\begin{itemize}
    \item We propose FPMVO to fuse strand orientations from multiple views in order to optimize them simultaneously. This accelerates outer directional field optimization by over 28\texttimes{}, while preserving accuracy.
    \item We develop PHG, a GPU-based parallel hair growing algorithm that simultaneously grows strands within a shared occupancy–orientation volume, achieving high geometric detail and curvature continuity with greatly improved accuracy and efficiency.
    \item Under a hybrid implicit–explicit framework, our system achieves an overall nearly 7\texttimes{} speed-up compared to existing state-of-the-art methods, while maintaining strong generalization and stability across different real-world hairstyles—including challenging curly hair—with comparable reconstruction quality.
\end{itemize}

\section{Related Research}

\subsection{Explicit Geometry Optimization Methods}
Pioneering work by Paris et al.~\cite{paris2004capture} first extracted strand orientations and geometric constraints from multi-view images for image-based reconstruction.
Subsequent research by Luo et al.~\cite{luoMultiviewHairCapture2012,luoStructureawareHairCapture2013,luoWideBaselineHairCapture2013} established a comprehensive multi-view capture framework integrating directional field estimation, structure-aware refinement, and wide-baseline optimization to achieve accurate reconstruction with fewer cameras.
Hu et al.~\cite{huRobustHairCapture2014} further proposed a framework based on a simulated example database learning structural priors and sketch-based retrieval mechanism, enabling single-view 3D modeling~\cite{maSingleViewHairModeling}.
Meanwhile, several methods explored alternative representations and automation. HairMeshes~\cite{yukselHairMeshes2009} introduced a mesh-based intermediate representation for improved structural consistency, while AutoHair~\cite{chaiAutoHairFullyAutomatic2016} automatically recovered strand orientation fields from a single image.
Zhang et al.~\cite{zhangDatadrivenApproachFourview2017} proposed a data-driven four-view reconstruction framework that jointly optimizes geometry and orientation.
Although these explicit methods achieved high geometric fidelity, they typically depend on dense camera setups, manual calibration, and global optimization, resulting in high computational cost and poor scalability.

\subsection{Implicit Neural Methods}
Saito et al.~\cite{saito3DHairSynthesis2018} introduced a volumetric variational autoencoder for 3D hair generation, while HairNet~\cite{zhouHairNetSingleViewHair2018} first predicted strand flow fields from a single image. 
Subsequent works~\cite{yangDynamicHairModeling2019,shenDeepSketchHairDeepSketchbased2019} incorporated temporal and sketch-conditioned constraints for dynamic or semantic modeling. 
Later, differentiable rendering frameworks such as Neural Strands~\cite{rosuNeuralStrandsLearning2022}, NeuralHDHair~\cite{wuNeuralHDHairAutomaticHighfidelity2022}, and DeepMVSHair~\cite{kuangDeepMVSHairDeepHair2022} advanced implicit volumetric modeling and multi-view integration for strand-level reconstruction. 
Diffusion- and domain-adaptation-based approaches~\cite{rosuDiffLocksGenerating3D,zhengHairStepTransferSynthetic2023} have further improved realism and generalization. 
Recent generative methods~\cite{hePermParametricRepresentation2024,zhengUnified3DHair2024,sunStrandHeadTextStrandDisentangled2024,liSimAvatarSimulationReadyAvatars2024} extend these ideas to multi-style, text-driven, and simulation-ready avatars. 
Despite significant progress in automation and generalization, most latent-based models still struggle to recover geometrically accurate and spatially aligned strand structures, especially for complex curly hairstyles.

\subsection{Hybrid High-fidelity Modeling Methods}
Recent research has increasingly focused on integrating explicit geometry with implicit neural representations, leading to unified high-fidelity frameworks for hair reconstruction.
Neural Haircut~\cite{sklyarovaNeuralHaircutPriorGuided2023} first combined strand-level priors with differentiable explicit geometry to enable neural reconstruction with improved controllability.
GaussianHair~\cite{luoGaussianHairHairModeling2024} unified geometric and appearance modeling through illumination-aware Gaussian fields, while GroomCap~\cite{zhouGroomCapHighFidelityPriorFree2024} realized multi-view high-fidelity hair capture without relying on pre-defined priors.
GaussianHaircut~\cite{zakharovHumanHairReconstruction2024} further introduced Gaussian primitive representations, achieving strand-aligned rendering at high precision with an unconstrained capture setup.

Although these hybrid methods have achieved remarkable progress in fidelity, they still suffer from high computational optimization overhead and frequent failures in reconstructing complex curly hairstyles.
MonoHair~\cite{wuMonoHairHighFidelityHair2024} takes an important step forward in this direction by combining multi-view directional optimization with implicit neural inference of internal structures, enabling strand-level hair reconstruction from monocular video. However, its performance is still constrained by sequential view-wise optimization and serial strand growing, which limit scalability and runtime efficiency.
Building upon these studies, our proposed EfficientMonoHair further advances the efficiency of hybrid explicit–implicit reconstruction through a unified multi-view direction fusion and parallel hair growing framework.

\section{Method}
\begin{figure*}[t]
  \centering
    \vspace*{-3mm}
    \captionsetup{skip=4pt}
    \includegraphics[width=\linewidth]{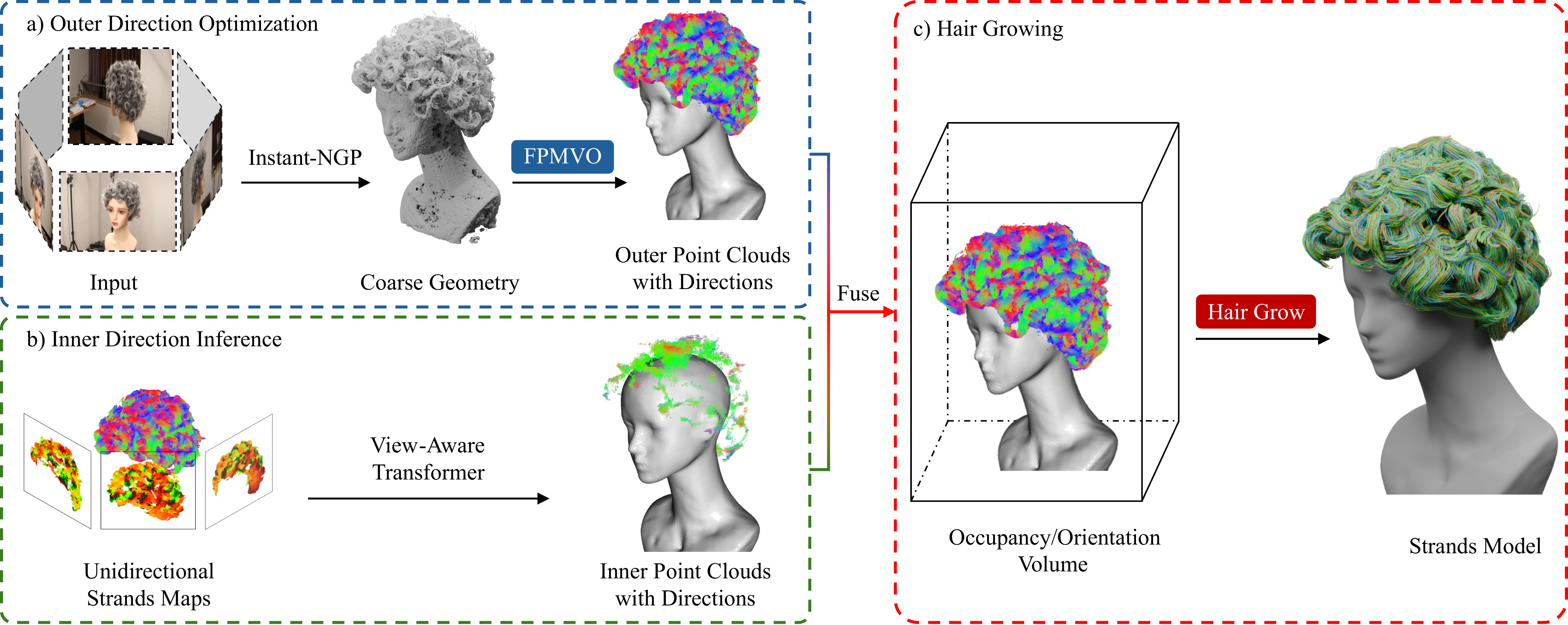}
    \caption{Our framework reconstructs strand-level hair geometry from a monocular video in three stages: (a) Outer Direction Optimization, where coarse hair geometry is reconstructed via Instant-NGP, followed by our FPMVO (detailed illustration in appendix) to obtain a stable outer point cloud with directions. (b) Inner Direction Inference, which leverages a View-Aware Transformer to infer invisible inner orientations from unidirectional strand maps; and (c) Hair Growing, where both inner and outer direction fields are fused into an Occupancy/Orientation Volume, from which our parallel strand growing (PHG) generates the final high-fidelity 3D hair model.}
   \label{fig: pipeline}
   \vspace{-2ex}
\end{figure*}

Our proposed \textbf{EfficientMonoHair} framework consists of three main stages: (a) Outer direction optimization, (b) Inner direction inference, and (c) Hair growing, as illustrated in Figure~\ref{fig: pipeline}.
\textbf{(a)} Given a monocular video input, we first employ Instant-NGP~\cite{muller2022instant} with converted COLMAP~\cite{schoenberger2016mvs,schoenberger2016sfm} cameras to reconstruct the scene. From this scene, we retrieve a coarse outer hair shape as a point cloud, denoted as $P_{\text{raw}}(\mathbf{p})$.
Subsequently, we use our Fusion-Patch-based Multi-View Optimization (\textbf{FPMVO}) to produce the outer-layer strand direction field. 
Through multi-view patch-level fusion, this module refines the outer directional information and produces a stable set of outer-layer direction-aware point clouds $P_{\text{out}}(\mathbf{p},\mathbf{d}_{\text{out}})$.
\textbf{(b)} To predict the internal regions of the hair volume, that are invisible in monocular video, we first render a undirectional strands map~\cite{wuMonoHairHighFidelityHair2024} and infer the internal orientations $P_{\text{in}}(\mathbf{p},\mathbf{d}_{\text{in}})$ using the View-Aware Transformer~\cite{wuMonoHairHighFidelityHair2024,kuangDeepMVSHairDeepHair2022}, similar to prior work.
\textbf{(c)} Finally, the inner and outer directional fields are fused into a unified occupancy and orientation volume using our Parallel Hair Growing (\textbf{PHG}) method. In contrast to prior approaches, PHG grows several strands of hair simultaneously, improving the robustness on low-accuracy orientation fields, while being about twice as fast.

\subsection{Fusion-Patch-based Multi-View Optimization}
Our method extends the sequential patch-based multi-view optimization introduced in MonoHair~\cite{wuMonoHairHighFidelityHair2024}. Similar to their approach, we first extract 2D orientation maps $\mathcal{O} \in \mathbb{R}^{H\times W\times 2}$ and confidence maps $\mathcal{C} \in [0,1]^{H\times W}$ using Gabor filters~\cite{feichtinger2012advances}.
The strand orientation this way tends to be sparsely distributed and inconsistent across different viewing angles. To address this, MonoHair~\cite{wuMonoHairHighFidelityHair2024} employs an exhaustive search strategy to iteratively determine the optimal orientation for each strand, which, however, incurs extremely high computational costs.  
In contrast to this, we propose the Fusion-Patch-based Multi-View Optimization (FPMVO) module, which efficiently \emph{fuses} directional consistency information across multiple views to avoid expensive iterations over the views, while maintaining stable and continuous direction fields.

In brief, FPMVO first performs (1) multi-view direction sampling and local orientation confidence estimation under each view.  
(2) We aggregate directional candidates over multiple views using medoid-based direction fusing to derive globally optimal strand directions $\mathbf{d}_{\text{out}}$. (3) Lastly, we enforce patch-based consistency on the fused directions to further refine and smooth the orientations. The result is an outer-layer direction-aware point cloud $P_{\text{out}}(\mathbf{p}, \mathbf{d}_{\text{out}})$ with high directional reliability.

\subsubsection{Multi-view Direction Sampling}
\label{sec: sampling}
To obtain stable strand orientation information from monocular video, we first generate a set of potential directional samples for each visible spatial point under all available views. To determine visibility, we compute a visibility score $V_i^v$ for each point $i$ and view $v$ by comparing the distance of $\mathbf{p}_i$ to $v$'s camera with this view's depth $\mathcal{D}_v(\mathbf{u}_i^v)$ at the projected location $\mathbf{u}_i^v = \Pi_v(\mathbf{p}_i)$. Here $\Pi_v(\cdot)$ denotes the projection function for the camera of view $v$. For each visible point $\mathbf{p}_i$ at each view  $v$, we can get the corresponding local 2D direction $\mathcal{O}_v(\mathbf{u}_i^v)$. We apply a pixel-wise directional offset on the image plane as
\begin{equation}
\mathbf{\check{u}}_i^{v} = \mathbf{u}_i^v + \lambda \, \mathcal{O}_v(\mathbf{u}_i^v),
\end{equation}
where $\lambda$ is a scalar controlling the offset magnitude.  
At the shifted position $\mathbf{\check{u}}_i^{v}$, we obtain the corresponding depth value $\check{z}_v = \mathcal{D}_v(\mathbf{\check{u}}_i^v)$ by sampling the rendered depth map.  
For each point, we generate a set of $S$ samples taken uniformly around the depth at the point to account for uncertainty in the depth estimation.
\begin{equation}
\check{z}_v^{(s)} = \check{z}_v + \delta^{(s)}, \quad \delta^{(s)} \in [-\Delta, \Delta].
\end{equation}
Note that the sample range is typically very narrow with $2\Delta = 10\text{mm}$. The shifted coordinates and sampled depths are then back-projected into world space using the inverse projection function $\Pi_v^{-1}(\cdot)$ to obtain candidate offset points in 3D.
\begin{equation}
\mathbf{\check{p}}_v^{(s)} = \Pi_v^{-1}(\mathbf{\check{u}}_i^{v}, \check{z}_v^{(s)}), \quad s \in \{1, \dots, S\}.
\end{equation}
Each of these candidate offset points then defines a direction when compared to the original point $\mathbf{p}_i$.
As a result of this process, we construct a set of multi-view candidate directions for each point $\mathbf{p}_i$, defined as 
\begin{equation*}
\mathbf{Dir}(\mathbf{p}_i) = \Biggl\{ \mathbf{d}_v^{(s)} = \frac{\mathbf{\check{p}}_v^{(s)} - \mathbf{p}_i}{\|\mathbf{\check{p}}_v^{(s)} - \mathbf{p}_i \|_2^2} \Biggr\}_{\substack{v \in \{1,\dots, V\}\\s \in \{1,\dots, S\}}}
\end{equation*}

\subsubsection{Medoid-based Direction Fusion}

After obtaining multi-view directional candidates for each point, the challenge lies in efficiently determining a globally consistent and stable direction field for the outer layer.  
Due to the variability of viewing angles, depth ambiguities, and potential occlusions, directly averaging candidate directions across views often leads to severe artifacts such as orientation blurring.  
To overcome this issue, we introduce a medoid-based direction fusion strategy, which aggregates geometrically consistent directional candidates in a robust manner to construct reliable strand orientations.

For each point $\mathbf{p}_i$, we collect the candidate direction set from the top-$K$ visible views based on confidence ranking:

\begin{equation}
\mathbf{Dir}_{tk}(\mathbf{p}_i) = \bigcup_{v \in \mathcal{V}_{tk}} \{ \mathbf{d}_{v}^{(1)}, \mathbf{d}_{v}^{(2)}, \dots, \mathbf{d}_{v}^{(S)} \},
\end{equation}
where $\mathcal{V}_{tk}$ denotes the subset of views with the highest $K$ confidence scores for each point (typically $K=5$).  
Each view $v$ provides $S$ directional samples $\mathbf{d}_{v}^{(s)}$ obtained from Sec~\ref{sec: sampling}, and the corresponding confidence weight $w_j$ is derived from the per-view confidence map $\mathcal{C}_v$.  
Thus, each point $\mathbf{p}_i$ has a total of $K \times S$ directional candidates across all views.

To determine the most consistent representative direction among these candidates, we first compute a pairwise similarity matrix within each depth layer:

\begin{equation}
G_{ij}^{(s)} = \big|\langle \mathbf{d}_{i}^{(s)}, \mathbf{d}_{j}^{(s)} \rangle\big|, \quad i,j \in \{1,\dots,K\},
\end{equation}
where $\langle \cdot, \cdot \rangle$ denotes the dot product between two unit direction vectors. The absolute value ensures bidirectional symmetry, since the hair strand segments do not have a unique forward or backward orientation.  
Based on this matrix, we define a weighted consistency score for each candidate direction as
\begin{equation}
\sigma_i^{(s)} = \frac{\sum_{j=1}^{K} w_j \, G_{ij}^{(s)}}{\sum_{j=1}^{K} w_j},
\end{equation}
This score reflects the geometric consistency of each directional candidate relative to all others within the same depth layer.  
Finally, we select the direction with the highest consistency score as the multi-view-fused orientation for at that given depth sample $s$:
\begin{equation}
\mathbf{d}_{f}^{(s)} = \mathbf{d}_{i^*}^{(s)}, \quad i^* = \arg\max_i \sigma_i^{(s)}.
\end{equation}
This process effectively finds the \emph{medoid direction} -- the direction with maximum weighted consistency on the unit sphere -- thereby preserving the geometric realism of local strand orientations while mitigating averaging artifacts.  
By repeating this operation across all $S$ depth samples, we obtain a compact set of globally fused directional candidates 
$\{ \mathbf{d}_f^{(1)} ,\dots, \mathbf{d}_f^{(S)} \}$ and its corresponding globally fused offset points $\{\mathbf{\check{p}}_f^{(1)} ,\dots, \mathbf{\check{p}}_f^{(S)}\}$.

\subsubsection{Patch-based Consistency Optimization}
After obtaining the globally fused offset points 
$\{\mathbf{\check{p}}_f^{(1)},\dots,\mathbf{\check{p}}_f^{(S)}\}$, we further decide the final direction through a 
patch-based consistency optimization. Unlike previous methods that perform multi-view weighted consistency, our fused candidate set is already view-invariant; thus, the optimization is performed directly in the image projection domain.
For each 3D point $\mathbf{p}_i$ and its fused candidate points 
$\{\mathbf{\check{p}}_f^{(1)} ,\dots, \mathbf{\check{p}}_f^{(S)}\}$, 
we first reproject them onto the image planes of the most confident views $\mathcal{V}_{tk}$ to obtain their 2D directional offsets:
\begin{equation}
\mathbf{r}_v^{(s)} =\Pi_v(\mathbf{\check{p}}_f^{(s)}) - \Pi_v(\mathbf{p}_i), v \in \mathcal{V}_{tk}
\label{eq:proj_r}
\end{equation} 
The set of 2D vectors 
$\mathbf{R} = \{\mathbf{r}_v^{(1)} ,\dots, \mathbf{r}_v^{(S)}\}$ 
captures the projected directional distribution of the fused candidates.
To evaluate their local consistency with the per-view orientation maps $\mathcal{O}_{v}$, 
we compute a bidirectional similarity score for each candidate within a small patch $\mathcal{O}_p^{\mathbf{u}_i^v} \in \mathbb{R}^{P\times P\times 2}$, that is centered around $\mathbf{u}_i^v$ with $p= 1,\dots, P^2$ serving as index into the patch:
\begin{equation}
\text{sim}_p^{(s)} = 
\max\big(
\langle \mathbf{r}_v^{(s)},\mathcal{O}_p^{\mathbf{u}_i^v} \rangle,
-\langle \mathbf{r}_v^{(s)},\mathcal{O}_p^{\mathbf{u}_i^v} \rangle
\big),
\label{eq:sim}
\end{equation}
where the symmetric formulation removes the ambiguity between
forward and backward strand orientations.
The corresponding local patch-wise consistency is defined as
\begin{equation}
\mathcal{L}_p^{(s)} = \text{sim}_p^{(s)}\mathcal{C}_p^{\mathbf{u}_i^v},
\label{eq:patchloss}
\end{equation}
where $\mathcal{C}_p^{\mathbf{u}_i^v}$ is the corresponding patch of the confidence map. We then select the maximal matching response within each patch:
\begin{equation}
\mathcal{L}_{\text{patch}}^{(s)} = \max_{p}\mathcal{L}_p^{(s)}.
\label{eq:patch_min}
\end{equation}
Finally, the optimal strand orientation is determined by selecting
the candidate with the highest patch consistency:
\begin{equation}
s^* = \arg\max_{s}\mathcal{L}_{\text{patch}}^{(s)},
\quad
\mathbf{d}_{\text{out}} = 
\frac{\mathbf{\check{p}}_f^{(s^*)} - \mathbf{p}_i}{\| \mathbf{\check{p}}_f^{(s^*)} - \mathbf{p}_i\|_2^2}.
\label{eq:final_dout}
\end{equation}

\subsection{Parallel Hair Growing}
\label{sec: hairgrowing}
To reconstruct full hair strands, we additionally require the internal hair structure $P_\text{in}(\mathbf{p}, \mathbf{d}_{in})$. To retrieve this, we follow MonoHair's multi-view transformer approach. We make a few performance improvements in this step, which we detail in the appendix.
After fusing the $P_{\text{in}}$ and  $P_{\text{out}}$, we obtain a unified volumetric occupancy–orientation representation 
$(\mathcal{V}_{\text{occ}}, \mathcal{V}_{\text{ori}})$
within 3D space. 

Starting from this input, the strand tracing method in MonoHair~\cite{wuMonoHairHighFidelityHair2024} relies on sequential CPU-based integration to iteratively advance each strand along the local orientation field. Each step depends on the result of the previous one, while global occupancy updates must be applied synchronously to avoid strand overlap. This design inherently limits parallelization, since the tracing of one strand cannot proceed until its predecessor has completed and the global voxel state has been updated. Moreover, the recursive segment-merging step requires iterative KD-tree rebuilding, further introducing strong serial dependencies and high memory overhead. These characteristics make the original algorithm difficult to scale to dense strand sets or high-resolution volumetric fields.

To overcome these limitations, we reformulate the strand-growing process as a fully parallelized volumetric tracing and linking pipeline that we call Parallel Hair Growing (\textbf{PHG}). 
Following the overall pipeline trace method of previous work~\cite{wuMonoHairHighFidelityHair2024,kuangDeepMVSHairDeepHair2022,wuNeuralHDHairAutomaticHighfidelity2022}, our PHG has two major improvements:
(1) Parallel guide strand initialization from scalp seeds and (2) Parallel segment growth and connection.

\begin{figure*}[thb]
    \centering
    \vspace*{-2mm}
    \captionsetup{skip=4pt}
    \includegraphics[width=\linewidth]
    {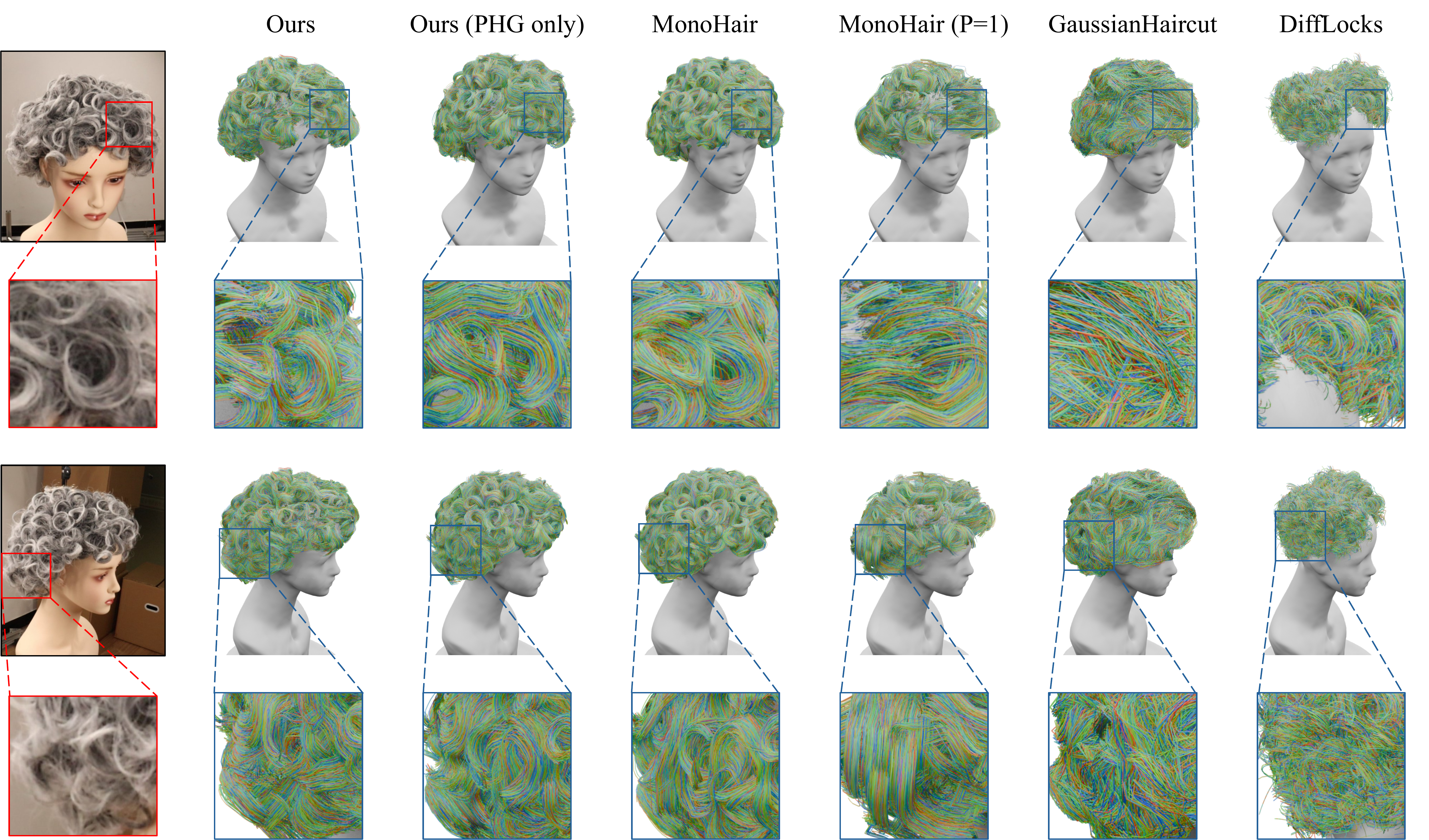}
    \caption{Visual comparison of the reconstruction of curly hair by multiple methods.}
    \label{fig:qualitative-curly}
    \vspace{-3mm}
\end{figure*}

\subsubsection{Guide Strand Initialization}
At the first stage, during the strand growth step, the original pipeline sets a global occupancy flag, ensuring that not too many strands occupy the same voxel during tracing.
While this guarantees strict spatial exclusivity, it also introduces hard synchronization dependencies that prevent parallel execution.
In contrast, our parallel formulation grows many strands in parallel and postpones the occupancy flag updates until the end of each batch tracing step. This relaxation allows multiple strands to grow simultaneously within the same local voxel region, effectively removing synchronization barriers and enabling fully batched SIMD-style tracing. Specifically, for each scalp point $\mathbf{h}_i$ with normal direction $\mathbf{n}_i$, we simultaneously trace multiple growth paths along $\mathbf{n}_i$ within $\mathcal{V}_{\text{ori}}$, forming a large set of guide strands $\mathcal{S}_{\text{root}}$. Tens of thousands of scalp seeds are traced concurrently, yielding efficient strand initialization with a total count of $N_{\text{root}}$. Although this strategy may temporarily allow multiple strand candidates to occupy overlapping voxels, the subsequent segment connection and merging stage naturally resolves these conflicts, since we apply interpolation to blend neighboring trajectories.
Furthermore, an ablation study in Sec.~\ref{sec: ablation-PHG} shows that this modification can improve the robustness of the hair growing, since we allow more strands, which can overcome inaccuracies in the orientation volume.

\subsubsection{Segment Growth and Connection}
After initial tracing, we introduce an ultra-fast spatial linking strategy 
to connect short strand segments into long, continuous strands. 
Compared to the original direction-based local merging in MonoHair~\cite{wuMonoHairHighFidelityHair2024}, our approach performs batched nearest-neighbor queries by constructing global KD-trees for all strand endpoints and performing vectorized batched nearest-neighbor queries to find compatible segment pairs 
that satisfy both spatial proximity and directional similarity constraints:
\begin{equation}
\|\mathbf{p}_i^{\text{end}} - \mathbf{p}_j^{\text{start}}\| < \delta_d, 
\qquad
\langle \mathbf{d}_i, \mathbf{d}_j \rangle > \cos(\delta_\theta),
\label{eq:segment_connection}
\end{equation}
where $\mathbf{p}_i^{\text{end}}$ and $\mathbf{p}_j^{\text{start}}$ denote 
the 3D endpoints of two candidate segments, 
and $\mathbf{d}_i$ and $\mathbf{d}_j$ represent their local tangent directions, respectively. 
The first condition enforces spatial proximity within a distance threshold $\delta_d$, 
while the second ensures angular consistency between the tangent directions 
within an angular tolerance $\delta_\theta$. 
Segments that meet both criteria are considered geometrically compatible 
and are merged into the same strand. 

Lastly, after connecting the individual segments, we construct a KD-tree on the scalp mesh $\Omega$ 
and project all unattached strand endpoints to their nearest scalp points sequentially, similar to MonoHair~\cite{wuMonoHairHighFidelityHair2024}. The overall strand generation process achieves 2–3$\times$ computational reduction compared to the state-of-the-art~\cite{wuMonoHairHighFidelityHair2024}, 
while preserving fine-level strand connectivity and shape fidelity.

\begin{table*}[t]
    \centering
    \vspace*{-3mm}
    \caption{\textbf{Quantitative Results on Hair20K}. Our method is competitive with MonoHair at patch size $P=5$, while being 6\texttimes{} faster. MonoHair's fast variant ($P=1$) is still considerably slower than Ours, while singificanlty degrading the accuracy. Diffusion based methods like DiffLocks are much faster, but produce comparatively inaccurate results.}
    \vspace{-2mm}
    \resizebox{\linewidth}{!}{
    \begin{tabu}{lrrccccccccccccccccccc}
    \toprule
    &  & 
    & \multicolumn{9}{c}{Occupation Accuracy (Voxel size in mm)} 
    & \multicolumn{9}{c}{Orientation Accuracy (Voxel size in mm / deg)} 
    \\
    \cmidrule(lr){4-12}
    \cmidrule(lr){13-21}
    & & 
    & \multicolumn{3}{c}{Precision} 
    & \multicolumn{3}{c}{Recall} 
    & \multicolumn{3}{c}{F1} 
    & \multicolumn{3}{c}{Precision} 
    & \multicolumn{3}{c}{Recall} 
    & \multicolumn{3}{c}{F1} 
    \\
    \cmidrule(lr){4-6}
    \cmidrule(lr){7-9}
    \cmidrule(lr){10-12}
    \cmidrule(lr){13-15}
    \cmidrule(lr){16-18}
    \cmidrule(lr){19-21}
    Method & Speed & Time
    & 2 & 3 & 4 & 2 & 3 & 4 & 2 & 3 & 4  
    & 2 / 20 & 3 / 30 & 4 / 40 & 2 / 20 & 3 / 30 & 4 / 40 & 2 / 20 & 3 / 30 & 4 / 40
    \\
    \midrule
    MonoHair~\cite{wuMonoHairHighFidelityHair2024} (P=5) & 1 \texttimes &136m 
    & 41.2 & 45.8 & 49.7 & \underline{73.2} & \underline{78.8} & \underline{82.4} & 50.6 & 56.3 & 60.6
    & \underline{21.1} & \underline{30.0} & \underline{38.7} & \textbf{22.5} & \textbf{29.4} & \textbf{36.1} & \underline{19.9} & \underline{27.3} & \underline{34.7}
    \\
    MonoHair~\cite{wuMonoHairHighFidelityHair2024} (P=1) & 1.2 \texttimes & 111m 
    & 37.3 & 41.7 & 45.4 & 63.0 & 69.7 & 74.2 & 44.8 & 50.4 & 54.7
    & 18.1 & 26.8 & 35.2 & 15.2 & 21.3 & 26.9 & 15.3 & 22.2 & 28.7
    \\
    Ours (full) & \textbf{5.9 \texttimes} & \textbf{23m}
    & \underline{45.0} & \underline{49.8} & \underline{53.8} & 68.7 & 75.6 & 80.0 & \underline{52.2} & \underline{58.4} & \underline{62.9} 
    & 19.0 & 27.4 & 38.3 & 15.0 & 20.2 & 27.6 & 15.5 & 21.7 & 30.3
    \\
    Ours (PHG only) & \underline{1.9 \texttimes} & \underline{72m}
    & \textbf{45.7} & \textbf{50.8} & \textbf{55.0} & \textbf{74.6} & \textbf{80.0} & \textbf{83.4} & \textbf{55.6} & \textbf{61.2} & \textbf{65.4}
    & \textbf{22.6} & \textbf{32.4} & \textbf{42.8} & \underline{19.8} & \underline{26.1} & \underline{33.2} & \textbf{20.5} & \textbf{28.2} & \textbf{36.5}
    \\
    \rowfont{\color{gray}}
    Ours (FPMVO only) & 0.9 \texttimes & 153m
    & 44.5 & 49.9 & 54.3 & 50.3 & 55.9 & 60.1 & 43.4 & 49.2 & 53.9
    & 15.5 & 25.5 & 37.8 &  8.9 & 13.3 & 19.7 & 10.1 & 15.9 & 23.8
    \\
    \rowfont{\color{gray}}
    DiffLocks~\cite{rosuDiffLocksGenerating3D} & 818 \texttimes & 10s 
    & 24.8 & 28.3 & 31.2 & 63.1 & 67.9 & 71.5 & 35.2 & 39.6 & 43.2
    &  8.2 & 15.6 & 22.4 & 13.6 & 21.2 & 26.8 & 10.0 & 17.6 & 24.0
    \\
    \bottomrule
    \end{tabu}
    } 
    \label{tab:metrics}
    \vspace{-3mm}
\end{table*}

\section{Experiments}
We evaluate our method on both real and synthetic datasets to comprehensively assess its performance. 
For the real data evaluation, we adopt multi-view hair video sequences from MonoHair~\cite{wuMonoHairHighFidelityHair2024}, 
which include representative hairstyles such as short, long, and curly hair, enabling visual comparisons under various appearance conditions. 

Since real hair datasets lack accurate strand-level ground truth geometry, we further conduct a quantitative evaluation on synthetic data. 
Specifically, we used the dataset Hair20K~\cite{Hair20k} extended from the USC-HairSalon~\cite{maSingleViewHairModeling}. 
Each sample is rendered from multiple viewpoints in Blender to provide multi-view images.

We mainly compare our EfficientMonoHair to two representative state-of-the-art approaches: 
The multi-view optimization method MonoHair~\cite{wuMonoHairHighFidelityHair2024} and the single-view neural implicit generation method DiffLocks~\cite{rosuDiffLocksGenerating3D}. 
MonoHair and DiffLocks, respectively, represent the explicit optimization and implicit neural generation paradigms.
All experiments for MonoHair and our method are conducted on an RTX 4090 GPU, while we evaluate DiffLocks on an A100 GPU using the official implementation. 

\subsection{Qualitative Evaluation}
\begin{figure}[t]
    \centering
    \vspace{-2mm}
    \captionsetup{skip=4pt}
    \includegraphics[width=\linewidth]{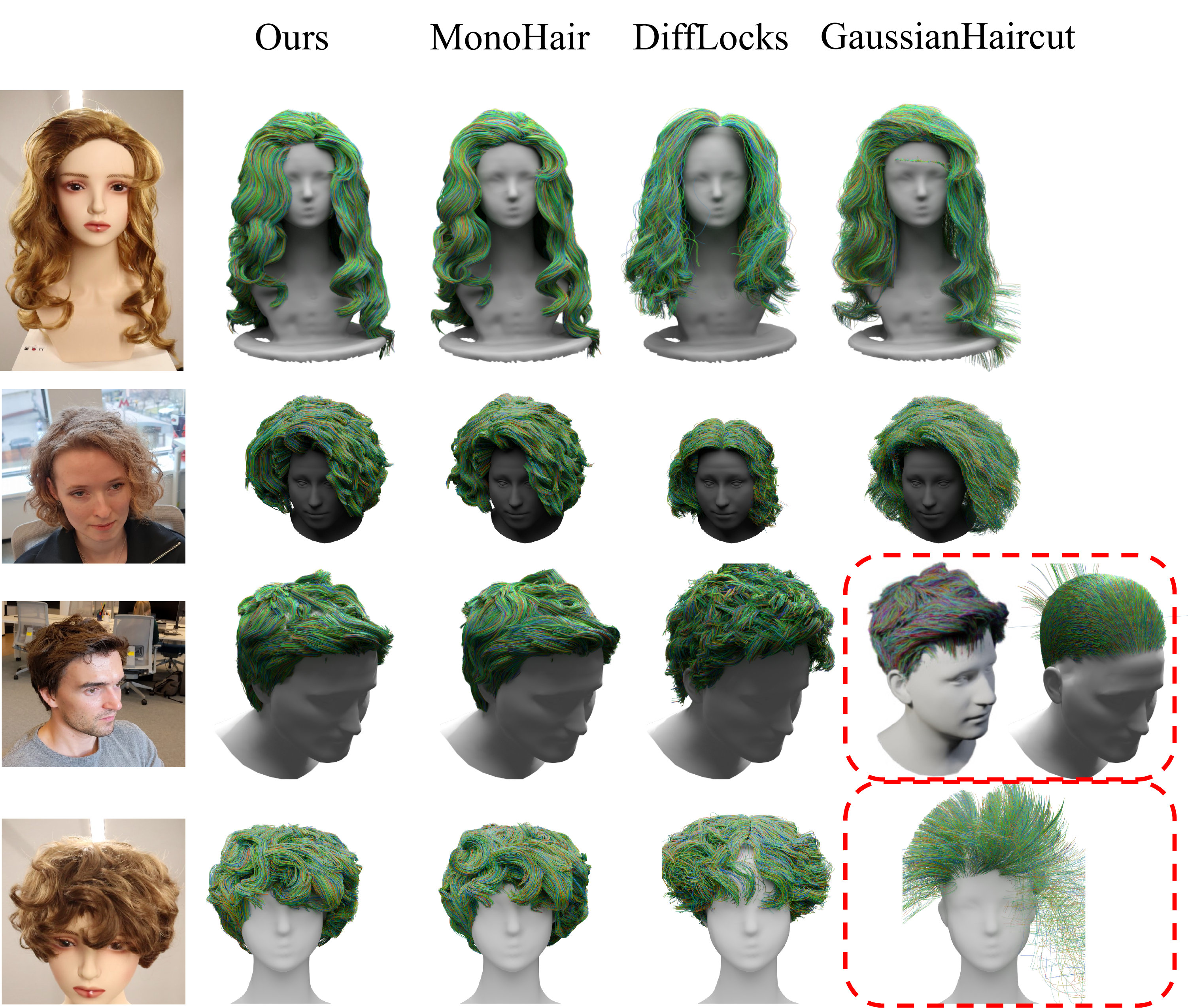}
    \caption{Visual comparison of the reconstruction of big wavy, asymmetric, short and medium curly hair styles by multiple methods. At the third row, we failed in reproducing this short hair using the GaussianHaircut codebase, therefore we present the results of Figure 7 in their paper~\cite{zakharovHumanHairReconstruction2024} next to our reproduced result. For the last row, GaussianHaircut also failed in the reconstruction. This shows the low robustness of their method compared to ours. }
    \label{fig:qualitative-other}
    \vspace{-3mm}
\end{figure}

We first conduct a qualitative visual comparison on real hairstyle video sequences, as shown in Fig.~\ref{fig:qualitative-curly} and Fig.~\ref{fig:qualitative-other}. 
The results demonstrate that EfficientMonoHair achieves strand-level detail preservation and directional continuity 
comparable to MonoHair. 
In particular, for highly complex hairstyles such as curly hair, our method effectively reconstructs intricate strand flow 
and a layered structure with a clear spatial hierarchy. 
In contrast, DiffLocks exhibits severe inconsistency in fine-scale strand orientations. 
Moreover, even the recent multi-view reconstruction approach, GaussianHaircut~\cite{zakharovHumanHairReconstruction2024}, fails to capture the detailed curvature of curly strands, 
further highlighting the robustness of our method in recovering complex hair geometries. Figure~\ref{fig:teaser} demonstrates that our method has sufficient quality to be used for convincing hair simulation. We show more hairstyles in the supplementary material.

\subsection{Quantitative Evaluation}
We also compare our method quantitatively to the state-of-the-art using synthetic data to measure occupancy and orientation precision, recall and F1. The occupancy metrics measure whether the reconstructed strands fall within the spatial neighborhood of the ground-truth geometry, 
while the orientation metrics jointly compare spatial proximity and directional consistency against different thresholds.

\subsubsection{Reconstruction Quality and Speed}
\begin{figure}[!tb]
    \centering
    \vspace{-2mm}
    \captionsetup{skip=8pt}
    \includegraphics[width=\linewidth]{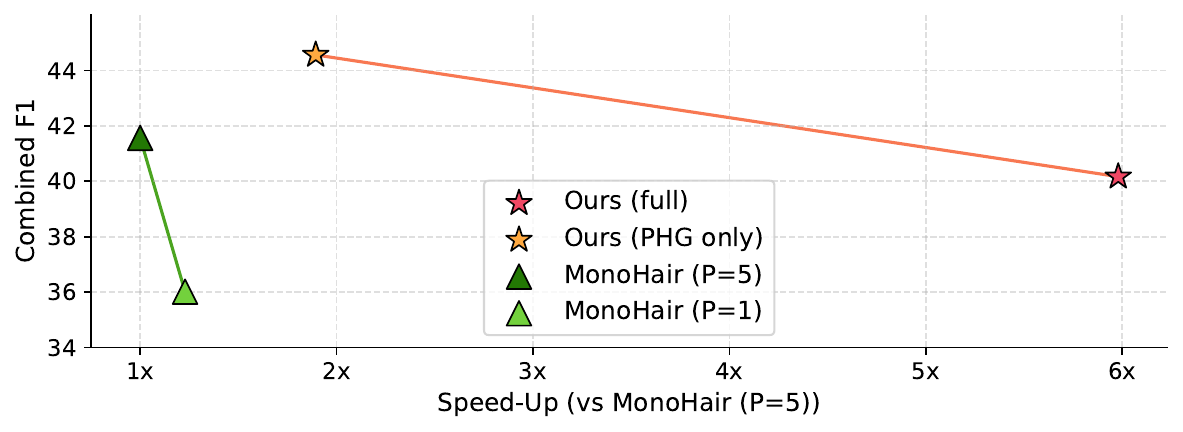}
    \caption{\textbf{Quality vs. Speed} using an average of the F1 scores for occupation and orientation on Hair20K.}
    \label{fig:speed-vs-f1}
    \vspace{-3mm}
\end{figure}

The results are shown in Table~\ref{tab:metrics} and we compare to MonoHair for different patch sizes in their patch-based multi-view optimization ($P=1$ and $P=5$), to compare both the highest quality and the fastest configuration for MonoHair. Note that our method always uses $P=5$ in FPMVO.
Our approach achieves approximately 6\texttimes{} faster reconstruction speed compared to MonoHair (P=5), while producing comparable strand geometry quality. Notably, our method even surpasses MonoHair in the occupancy metrics, indicating improved spatial completeness, while achieving up to 88\% of its orientation F1 score. Moreover, our method consistently outperforms DiffLocks across all metrics.
Compared to MonoHair's fast version (P=1), which degrades significantly in quality for just a slight speed-up, we demonstrate that our method provides a superior accuracy / speed trade-off. We illustrate this trade-off in Figure~\ref{fig:speed-vs-f1} where we compare the combined F1 score of occupancy and orientation with the speed-up of the different methods.

\subsubsection{Effect of FPMVO and PHG}
\label{sec: ablation-PHG}
We can see that our individual contributions have different effects on the results. To investigate this in detail, we test both contributions in isolation: \textit{Ours (FPMVO only)} uses FPMVO for initial orientations and MonoHair's hair growing strategy. \textit{Ours (PHG only)} uses MonoHair's unfused PMVO for initial orientations with PHG.

As shown in Table~\ref{tab:metrics}, \textit{Ours (PHG only)}, the combination of MonoHair's PMVO and our PHG, achieves the most accurate reconstruction, both in terms of 
occupancy and orientation. This indicates that MonoHair's original PMVO achieves more accurate initial orientations compared to FPMVO, which are then complemented by the robustness of our PHG strategy. In contrast, \textit{Ours (FPMVO only)} uses MonoHair's non-robust growing strategy, leading to significant degradation. While the slightly lower initial orientation accuracy of FPMVO leads to failure with the non-robust hair growing, we see that in \textit{Ours (full)}, PHG produces an accurate reconstruction regardless. This demonstrates that our PHG strategy substantially enhances the robustness of the system to direction-field noise. As illustrated in Fig.~\ref{fig:qualitative-curly}, 
the \textit{Ours (PHG only)} configuration also restores finer strand details on real data compared with the baseline. 

From an algorithmic perspective, MonoHair's hair growing performs step-wise occupancy checking to avoid collisions with existing strands during growth (as discussed in Sec.~\ref {sec: hairgrowing}). This strict constraint guarantees geometric consistency,
but amplifies the impact of local direction errors. In contrast, our proposed PHG adopts batched strand generation and local occupancy masking to perform spatial filling and connection in parallel, alleviating local conflict constraints and improving tolerance to directional noise. This design allows PHG to maintain macroscopic geometric continuity while significantly accelerating the reconstruction process and preserving plausible strand morphology even under imperfect direction fields.

\begin{figure}[t]
    \centering
    \captionsetup{skip=4pt}
    \includegraphics[width=\linewidth]{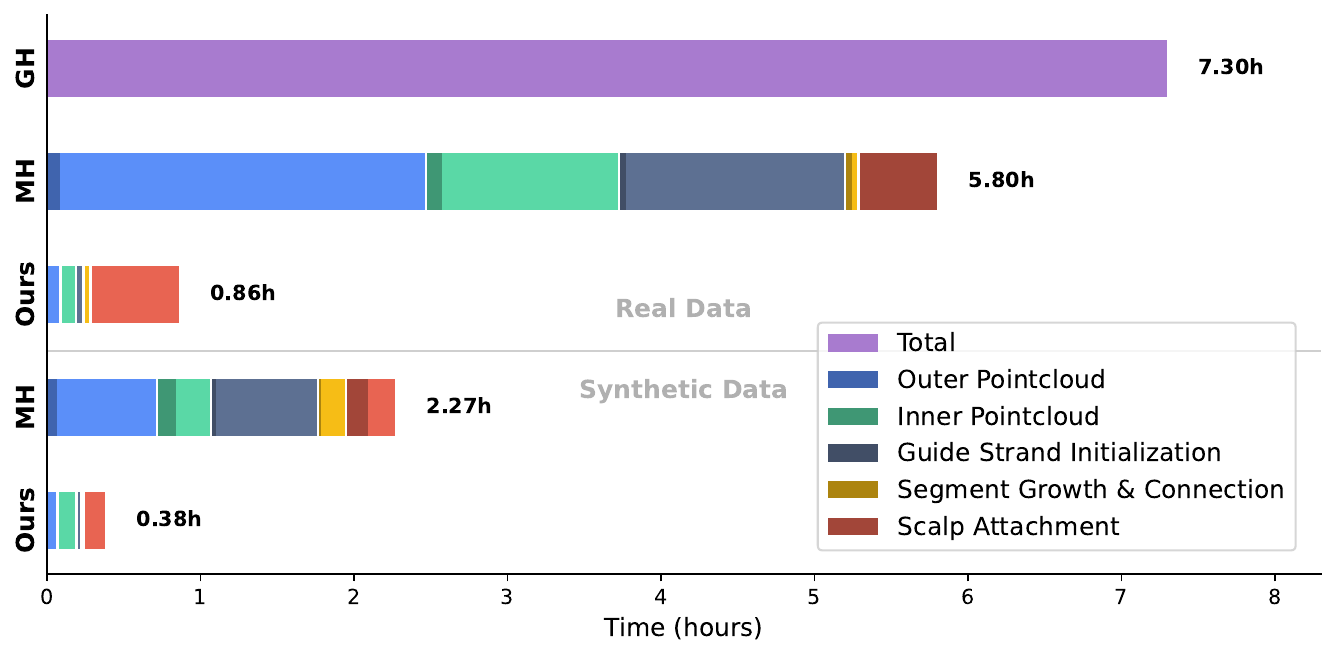}
    \caption{\textbf{Time Breakdown.} We compare the timing of our method to MonoHair (MH) and GaussianHaircut (GH), split into the individual steps of the pipeline where applicable. Darkened regions on top indicate our step time. Results are averaged over seven sequences of the real captured data shown in this paper.}
    \label{fig:time-breakdown}
    \vspace{-3mm}
\end{figure}

\subsubsection{Acceleration of Individual Components}
To verify the contribution of each design component to the overall efficiency improvement, we conduct a stage-wise time ablation analysis of the EfficientMonoHair framework, as illustrated in Fig.~\ref{fig:time-breakdown}. 
By comparing each processing stage with its counterpart in MonoHair's pipeline, we quantitatively evaluate the computational acceleration achieved by our proposed strategies across three major stages: Outer Point Cloud Optimization, Inner Point Cloud Inference, and Parallel Hair Growing, which includes three steps.
We measure the running time of each step on both real data and synthetic data. Due to the simpler structure and lower strand density of synthetic hairstyles, the overall reconstruction time is significantly shorter than that of real data. On real data, our method achieves substantial acceleration in all stages except the final Scalp Attachment step, with the most pronounced improvement observed in the Outer PointCloud Optimization phase (FPMVO) with a 28\texttimes{} speed-up.



\section{Limitation and Future Work}
We inherit MonoHair's limitation and our framework still struggles to reconstruct highly entangled hairstyles, such as braids, buns/updos, and tightly tied hairstyles. Incorporating stronger geometric or physical priors to explicitly model inter-strand dependencies 
would be an interesting direction for future research. 
Although our parallel hair-growing algorithm substantially improves efficiency, the last stage of connecting strands to the scalp still requires sequential processing for each strand. This limitation arises from the KD-tree–based nearest-neighbor search used for matching strand endpoints to scalp roots. Developing a more parallelization-friendly strategy for large-scale strand-to-scalp attachment represents a promising avenue for future work.
\section{Conclusion}

We present EfficientMonoHair, a fast and unified framework for strand-level hair reconstruction from monocular video. 
Under an implicit–explicit hybrid optimization, our method achieves efficient coordination between direction optimization and strand growing. The proposed FPMVO module enables rapid multi-view direction aggregation, thereby significantly reducing the number of optimization iterations. Meanwhile, our PHG module adopts a parallel strand-growing strategy that performs robust large-scale, synchronized strand generation and connection within a unified occupancy–orientation volume.
Experimental results demonstrate that EfficientMonoHair achieves a 6–8$\times$ acceleration 
over the existing state-of-the-art, while maintaining strand-level geometric detail and directional consistency. Using just our PHG strategy yields the most accurate strand-level hair reconstructions to date with a 2\texttimes{} speed-up over the prior state-of-the-art.
Our ablation studies confirm the complementarity effect of our contributions: FPMVO improves directional fusion efficiency, and PHG enhances robustness to noisy direction fields. 
Overall, our framework achieves a strong balance between reconstruction accuracy and computational efficiency, 
offering a new solution for high-fidelity strand modeling from monocular videos.

\section*{Acknowledgements}
This work was supported by the King Abdullah University of Science and Technology (KAUST)  research fund under grant BAS/1/1680-01-01.

\printbibliography
\end{document}